\documentclass{ifacconf}
\usepackage{stfloats}
\usepackage{graphicx} 
\usepackage{caption} 
\usepackage{indentfirst} 
\usepackage{xcolor}
\usepackage{amsmath}
\usepackage{graphicx}      
\usepackage{natbib}        
\usepackage{float}
\usepackage{multirow}
\usepackage{tikz}
\usepackage{svg}
\usepackage{adjustbox}
\usepackage{subcaption}
\usepackage{breakurl}
\usepackage[hyphens]{url}

\begin{document}

\begin{frontmatter}

\title{Beyond ADE and FDE: A Comprehensive Evaluation Framework for Safety-Critical Prediction in Multi-Agent Autonomous Driving Scenarios}

\thanks[footnoteinfo]{Correspond Author: Zhijian He (e-mail: hezhijian@sztu.edu.cn)}

\author[First]{Feifei Liu} 
\author[First]{Haozhe Wang} 
\author[First]{Zejun Wei}
\author[First]{Qirong Lu}
\author[Second]{Yiyang Wen}
\author[First]{Xiaoyu Tang}
\author[Second]{Jingyan Jiang}
\author[Second]{Zhijian He}
\address[First]{South China Normal University, Shanwei, Guangdong, China 
(e-mail: 20238331056@m.scnu.edu.cn, 202481324245@m.scnu.edu.cn, 202481313613@m.scnu.edu.cn)}
\address[Second]{Shenzhen Technology University, Shenzhen, Guangdong, China
   (e-mail: jiangjingyan@sztu.edu.cn, 202200202091@stumail.sztu.edu.cn, hezhijian@sztu.edu.cn)}

\begin{abstract}                
Current evaluation methods for autonomous driving prediction models rely heavily on simplistic metrics such as Average Displacement Error (ADE) and Final Displacement Error (FDE). While these metrics offer basic performance assessments, they fail to capture the nuanced behavior of prediction modules under complex, interactive, and safety-critical driving scenarios. For instance, existing benchmarks do not distinguish the influence of nearby versus distant agents, nor systematically test model robustness across varying multi-agent interactions. This paper addresses this critical gap by proposing a novel testing framework that evaluates prediction performance under diverse scene structures, saying, map context, agent density and spatial distribution. Through extensive empirical analysis, we quantify the differential impact of agent proximity on target trajectory prediction and identify scenario-specific failure cases that are not exposed by traditional metrics. Our findings highlight key vulnerabilities in current state-of-the-art prediction models and demonstrate the importance of scenario-aware evaluation. The proposed framework lays the groundwork for rigorous, safety-driven prediction validation, contributing significantly to the identification of failure-prone corner cases and the development of robust, certifiable prediction systems for autonomous vehicles.
\end{abstract}


\end{frontmatter}
\section{Introduction}
\setlength{\parskip}{0pt}
Prediction is essential in autonomous driving by \cite{mfmos} and \cite{yu2025enduring}, which is integrated with semantic maps and agent behavior modeling. Semantic maps provide critical context, such as road structure and traffic signals, while agent prediction models anticipate the future trajectories of dynamic road users. This synergy enables the system to foresee potential conflicts and make proactive decisions, enhancing both safety and navigation efficiency. By forecasting future states, the vehicle can adapt to changing traffic conditions, avoid collisions, and optimize route planning. Therefore, predictive capabilities are fundamental for both the safety and reliability of autonomous systems, facilitating the progression towards fully autonomous driving.

The comprehensiveness of prediction in autonomous driving by~\cite{overleapmamba} is inherently difficult to define due to the limitations of current evaluation metrics, which predominantly rely on simple metrics such as Final Displacement Error (FDE) or Average Displacement Error (ADE) by~\cite{8578338} and \cite{yuan2021agentformer}. These metrics, while useful for assessing the overall accuracy of an algorithm, fail to capture the nuances of where and how the model performs across different scenarios. For example, these standards do not provide insight into the influence of nearby versus distant agents \cite{yuan2021agentformer}, making it unclear whether an algorithm's predictions are more sensitive to local interactions or broader contextual factors. A more critical issue lies in the inability of these testing frameworks to identify and address safe-critical scenarios where the system's performance is vital for safety by \cite{safety}. For example, consider an autonomous vehicle approaching an intersection where a pedestrian unexpectedly steps onto the crosswalk, or a vehicle in an adjacent lane suddenly swerves into the vehicle's path. These scenarios are particularly critical because misjudgments in predicting the behavior of the pedestrian or other vehicle could lead to accidents. However, current evaluation metrics such as FDE and ADE primarily focus on general accuracy and displacement errors, rather than the system's ability to correctly handle such high-risk situations. As a result, the lack of focus on these safe-critical scenarios often leads to misjudgments, undermining the reliability and robustness of prediction models in the most critical driving contexts. 

To summarize, the main contributions of this paper are:
(1) We propose a three-layer safety evaluation framework that tests prediction models across semantic information, agent density, and spatial distribution. (2) Complex driving environments are broken down into detailed classifications based on the filter framework. Comprehensive robustness evaluation becomes feasible, moving beyond traditional single-condition testing toward systematic multi-scenario analysis. (3) Complexity of environments is brought into focus through this scenario decomposition process, as a key evaluation dimension, filling the critical gap in understanding how environmental complexity affects prediction performance beyond simple trajectory accuracy metrics.
\section{Related Work}
\subsection{Sequence Modeling}
Sequence modeling forms the foundation for trajectory prediction in autonomous systems. Early Recurrent Neural Network  (RNN) architectures by \cite{1990Finding} were applied to capture temporal patterns in agent movements over time. Long Short-Term Memory (LSTM) by \cite{LSTM} addressed gradient vanishing issues and became widely adopted for sequential trajectory learning. Recently, Transformer-based methods by \cite{2021transformer}and \cite{yuan2021agentformer} have emerged for modeling long-range dependencies through self-attention mechanisms. Graph-enhanced sequence models by \cite{li2020evolvegraph} combined spatial relationships with temporal modeling for multi-agent scenarios. These approaches typically encode agent's historical states and decode future trajectories based on learned temporal patterns. However, sequence models perform poorly in scenarios without semantic map constraints, relying solely on movement history for prediction. More critically, existing sequence models show limited sensitivity to agent density changes and map information variations. This insensitivity prevents comprehensive assessment of model safety in dynamic scenarios. Our framework addresses these overlooked limitations.
\subsection{Trajectory Prediction}
Trajectory prediction has evolved from simple certainty methods like RNNs by~\cite{Ala:16}, social forces~\cite{PhysRevE.51.4282} and Gaussian processes~\cite{4359316} to more sophisticated approaches. Given the inherently uncertain and multi-modal nature of future trajectories, modern methods use probabilistic models including Conditional Variational Autoencoders (CVAEs), Generative Adversarial Networks (GANs) \cite{11027605}, diffusion models \cite{10748368}, and transformer architectures like AgentFormer \cite{yuan2021agentformer}. Safety-aware approaches for multi-vehicle scenarios have also emerged \cite{10726633}.Current methods rely heavily on attention mechanisms and social interaction modeling. However, they show limited robustness across different scenarios, particularly with varying agent density, spatial distribution, and semantic information. While achieving good performance on standard metrics like ADE and FDE, they lack consistent reliability in safety-critical conditions. The core issue lies not in the prediction algorithms themselves, but in traditional evaluation frameworks that fail to expose model vulnerabilities under diverse driving environments.

\subsection{Cyber Physical System Testing}
Traditional testing of cyber-physical systems, particularly in autonomous driving, often relies on black-box, full-input testing methods as discussed by \cite{dosovitskiy2017carla}. As pointed out by \cite{ji2025autonomous}, these methods involve manipulating configurable driving environments to trigger specific reactions from autonomous driving systems, with test cases specifying comprehensive configurations of vehicle movements and environmental elements. This process effectively implements a full computer-to-physical mapping. However, in real-world scenarios, autonomous driving systems cannot access complete input information, rendering full-input testing methods unrealistic and impractical. To address this gap, our framework better approximates real-world conditions by controlling the availability of input information and the complexity of the testing scenarios.
Furthermore, as highlighted by  \cite{dosovitskiy2017carla}, existing safety-critical testing approaches face significant computational overhead due to the nature of full-input black-box testing. A fundamental contradiction exists between the vastness of possible scenario spaces and the limitations imposed by computational resources. As a result, traditional black-box testing methods can only cover a small fraction of potential scenario combinations, leaving many safety-critical scenarios inadequately verified and undermining the completeness and reliability of testing.

To mitigate these challenges, our safety evaluation framework employs multi-dimensional screening to identify safety-critical scenarios, thus avoiding the substantial resource consumption associated with full-input black-box testing. In conclusion, our evaluation framework integrates scenario screening with controlled input information, operating under the constraints of limited sample conditions. This approach allows for the efficient identification and assessment of safety-critical scenarios in trajectory prediction systems, ultimately enhancing the reliability of autonomous driving system validation.

\begin{figure*}[t]
\centering
\begin{tabular}{cc}
\includegraphics[width=1.0\textwidth, height=1.0\textheight, keepaspectratio]{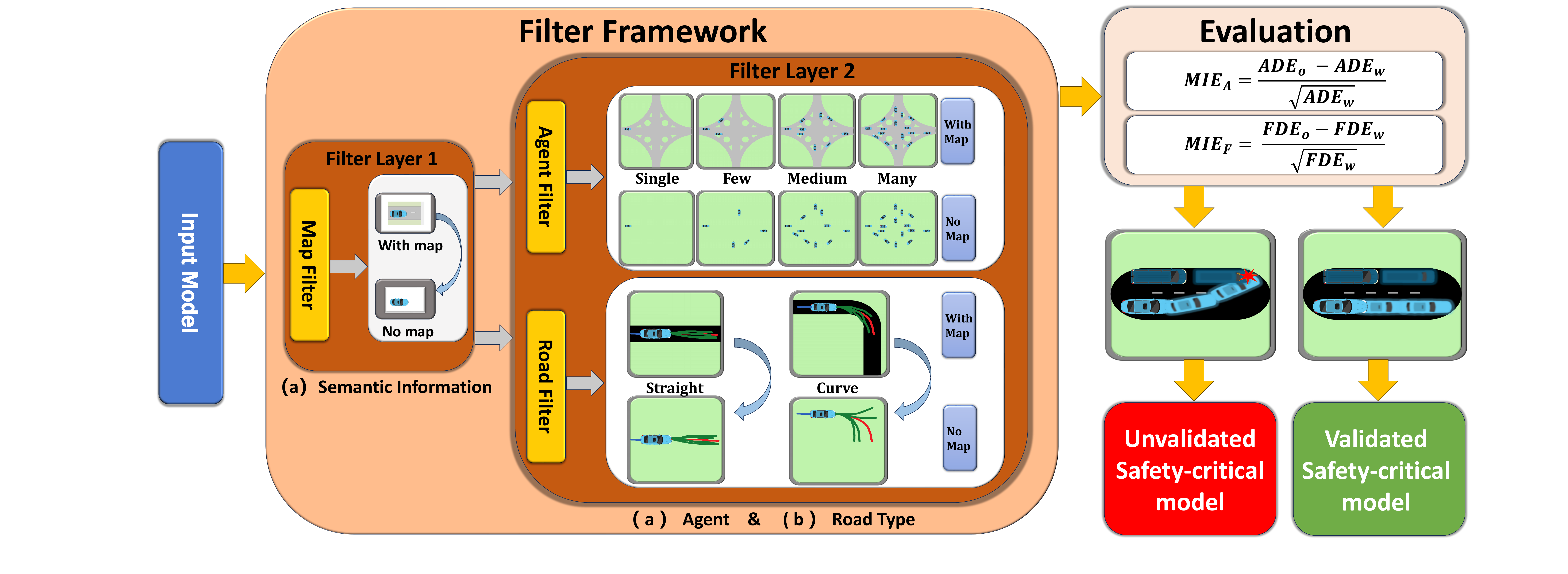} 
\end{tabular}
\vspace{0.2cm}
\caption{\textbf{Overview of the proposed three-layer safety-critical evaluation framework for trajectory prediction models.} The framework consists of layer 1(a) evaluating model performance with and without semantic maps using the proposed MIE metric, layer 2(b) assessing performance across four density levels, and layer 2(c) evaluating performance on straight versus curved road scenarios. Only models passing all three filters are certified as safety-critical for autonomous driving applications.}
\label{fig:method}
\end{figure*}

\section{METHODOLOGY}
%
\subsection{Framework Overview }
In this section, we propose a comprehensive evaluation framework for safety-critical prediction in multi-agent autonomous driving scenarios. Fig.~\ref{fig:method} illustrates the workflow of our safety evaluation framework. 
As shown in Fig.~\ref{fig:method}(a) Semantic Information, the first filter assesses the model's performance with and without semantic map information, a critical distinction that influences the model's ability to handle safety-critical predictions. The absence of semantic maps introduces a significant challenge, as these scenarios simulate real-world conditions where the model must operate with inherent uncertainty, devoid of prior knowledge about the road environment. 
Subsequently, as illustrated in Fig.~\ref{fig:method}(a) Agent, building on the semantic classification from Layer 1, the second filter further categorizes both with map and  without map scenarios by varying levels of agent density (single, few, medium, and many). This results in eight distinct scenario types that provide a nuanced understanding of how multi-agent interactions influence the model's safety-critical predictive performance under different informational contexts.
Finally, Fig.~\ref{fig:method}(b) Road Type shows that the third filter integrates road geometry classification (straight versus curved) with the semantic information from Layer 1. This classification creates distinct scenarios involving straight and curved roads under both with map and without map conditions. It aims to elucidate the interaction between geometric constraints and the availability of semantic information, shedding light on how these factors collectively affect the model's performance in safety-critical situations.
Only those models that demonstrate robust performance across all three filtering layers are considered validated for safety-critical applications in autonomous driving.
\subsection{Semantic Information Processing}
Initially, we handle the scenarios by both retaining and removing semantic information, thereby generating two distinct versions of each scenario, one incorporating semantic information and the other devoid of it. Existing autonomous driving datasets contain complex data, with semantic maps providing prediction models with crucial prior knowledge, such as road restrictions and behavioral rules. In the absence of this fundamental information, models must instead rely on historical trajectories and interactions among agents to make predictions. This shift fundamentally alters the nature of the prediction problem. In real-world applications, it is essential to account for situations where semantic maps are unavailable. Therefore, we propose that semantic information be considered a primary criterion in the safety evaluation framework, forming the basis for layered assessment.

\subsection{Multi-Dimensional Scenario Classification}
However, semantic information alone is insufficient for evaluation. Agent density further impacts model's map-dependency. Even well-performing models may frequently fail in single-agent scenarios. Prediction models face exponentially increasing computational demands when modeling dense multi-agent interactions. Moreover, prediction errors in high-density scenarios pose greater safety risks, as a single error can trigger cascading failures across multiple agents. 
 Similarly, different spatial distributions create distinct motion constraints and prediction challenges for vehicles. Therefore, we conduct safety assessments considering both agent density and spatial distribution across all scenarios. Specifically, we formalize our complete scenario classification framework as a hierarchical partitioning system :
\begin{equation}
S^{eval} = \bigcup_{\sigma\in\{\text{m},\text{n}\}} \bigcup_{\rho\in\{\text{sing},\text{few},\text{med},\text{many}\}} \bigcup_{\tau\in\{\text{str},\text{curv}\}} S_{\sigma,\rho,\tau}^{(M)}
\end{equation}
where each scenario subset is defined as:
\begin{equation}
S_{\sigma,\rho,\tau}^{(M)} = \{s \in S_{\text{r}} \mid \Gamma_{\text{s}}(s) = \sigma, \Phi_{\text{d}}(s) = \rho, \Psi_{\text{g}}(s) = \tau\}
\end{equation}

Here $\Gamma_{\text{s}}$, $\Phi_{\text{d}}$, and $\Psi_{\text{g}}$ represent the semantic, density, and geometric classification functions, respectively.
The parameters $\sigma$, $\rho$, and $\tau$ denote semantic completeness, agent density level, and geometric type, while $M$ means model. This split creates comprehensive scenarios, ensuring systematic classification for safety assessment.
\section{Experiments}
Traditional evaluation metrics like ADE and FDE provide valuable insights but fail to capture key real-world challenges. To address these limitations, we conduct three targeted experiments to evaluate model robustness in more complex scenarios. (1) With/Without Map Information Comparison. Most evaluations rely on map-based scenarios, assuming that semantic maps are available. However, in many real-world applications, maps are unavailable or unreliable. We assess model performance with and without map information to understand how the absence of maps impacts prediction accuracy, revealing potential vulnerabilities in environments without semantic maps. (2) Agent Density Comparison. Existing research typically examines models in scenarios with normal agent densities, but performance can vary significantly with agent density. In high-density environments, where accident risk and severity are higher, the model’s ability to predict safely becomes critical. Therefore, we evaluate model performance under different agent densities to assess how it handles safety-critical situations in crowded settings.
(3) Spatial Distribution Comparison. Traditional models perform well on straight roads, but prediction accuracy often decreases on curved roads, where trajectories are more complex. Since curved roads introduce specific safety risks, we differentiate between straight and curved road scenarios to assess how well the model manages these increased complexities, which are crucial for real-world safety. These experiments evaluating performance in environments without semantic maps, under varying agent densities, and across curved and straight road scenarios—ensure a more comprehensive safety assessment, identifying models that are robust enough for real-world deployment.

\textbf{Implementation Details.}
We evaluate on the nuScenes by \cite{caesar2020nuscenes} dataset using AgentFormer by \cite{yuan2021agentformer} as our baseline model, predicting 6-timestep (3s) future trajectories from 4-timestep (2s) observations with K=20 trajectory samples.
\subsection{With/Without Map Information Comparison}
\begin{figure}[tbp]
    \centering
    \includegraphics[width=1\columnwidth]{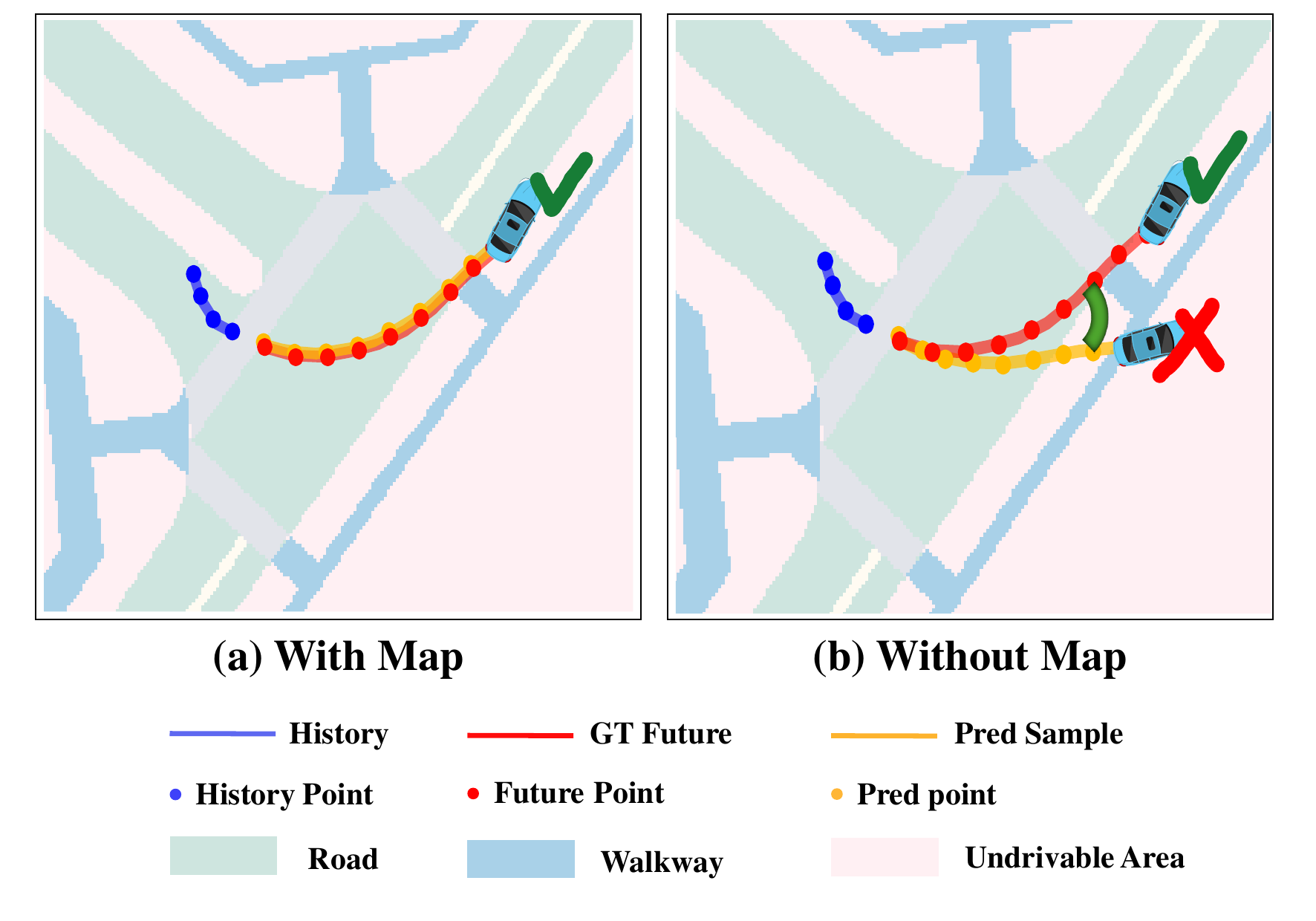}
    \caption{\textbf{Visualization of Trajectory Samples with and without Maps. } (a) with map scenario at an intersection. (b) without map scenario at an intersection.}
    \label{fig:curve_map_compairson}
\end{figure}
Current evaluation methods mainly focus on map-based scenarios and ignore model performance in map-free conditions. This creates a significant gap since many real-world areas lack map information, including construction zones, rural roads, and remote regions. To address this limitation, we compare prediction performance with and without semantic maps.
To quantify model's dependency on semantic map information, we introduce the Map Information Effectiveness (MIE) metric using square root normalization to address scale sensitivity:
\setlength\abovedisplayskip{4pt}
\setlength\belowdisplayskip{4pt}
\begin{equation}
\text{MIE} = \frac{{\text{Error}_{\text{o}}} - {\text{Error}_{\text{w}}}}{\sqrt{\text{Error}_{\text{o}}}} 
\end{equation}

Where \textbf{o} means \textbf{without map}, \textbf{w} means \textbf{with map}. Positive values indicate map dependency. The larger the value, the more the model depends on the semantic information of the map. We substitute the ADE and FDE values into Error to get $MIE_A$ and $MIE_F$, respectively.
In Table~\ref{tab:overall_mie}, we present the results showing significant map dependency across all scenarios. Specifically, for scenarios with maps, the model's ADE and FDE are 1.86 m and 3.89 m, respectively. In contrast, for scenarios without maps, the ADE and FDE increase to 1.98 m and 4.21 m, respectively. These results demonstrate that the model's performance is notably worse without semantic map information. Furthermore, the MIE metric reveals that the model's performance is heavily dependent on map information. With improvements of 0.08m in $MIE_A$ and 0.16m in $MIE_F$, the model performs substantially better when map is available. These findings indicate traditional metrics such as ADE and FDE cannot reveal this map-dependent performance variation, making MIE a more comprehensive metric for evaluating model robustness.

\begin{table}[tbp]
\centering
\caption{Overall Map Information Effectiveness}
\label{tab:overall_mie}
\begin{tabular}{lccccc}
\hline
Semantic & Sample & ADE & FDE & $MIE_A$ & $MIE_F$ \\
Information & Size & (m) & (m) & & \\
\hline
With Map    & 9041 & 1.8558 & 3.8892 & \multirow{2}{*}{0.0807} & \multirow{2}{*}{0.1622} \\
Without Map & 9041 & 1.9754 & 4.2051 & & \\
\hline
\end{tabular}

\end{table}

In Fig.~\ref{fig:curve_map_compairson}, we show that current prediction models exhibit a dangerous over-reliance on semantic map data, creating potential safety hazards. In scenarios without map, the models exhibit a much higher risk of misprediction, leading to unsafe outcomes. This is particularly critical for real-world driving, where map data is not always available. In such cases, the model's performance degradation becomes a significant safety concern, as erroneous predictions could lead to accidents. Therefore, only models that perform well in both scenarios (with and without maps) should be considered safe for deployment in autonomous driving applications. 
\subsection{Agent Density Comparison}

\begin{figure}[tbp]
    \centering
    \includegraphics[width=1\columnwidth]{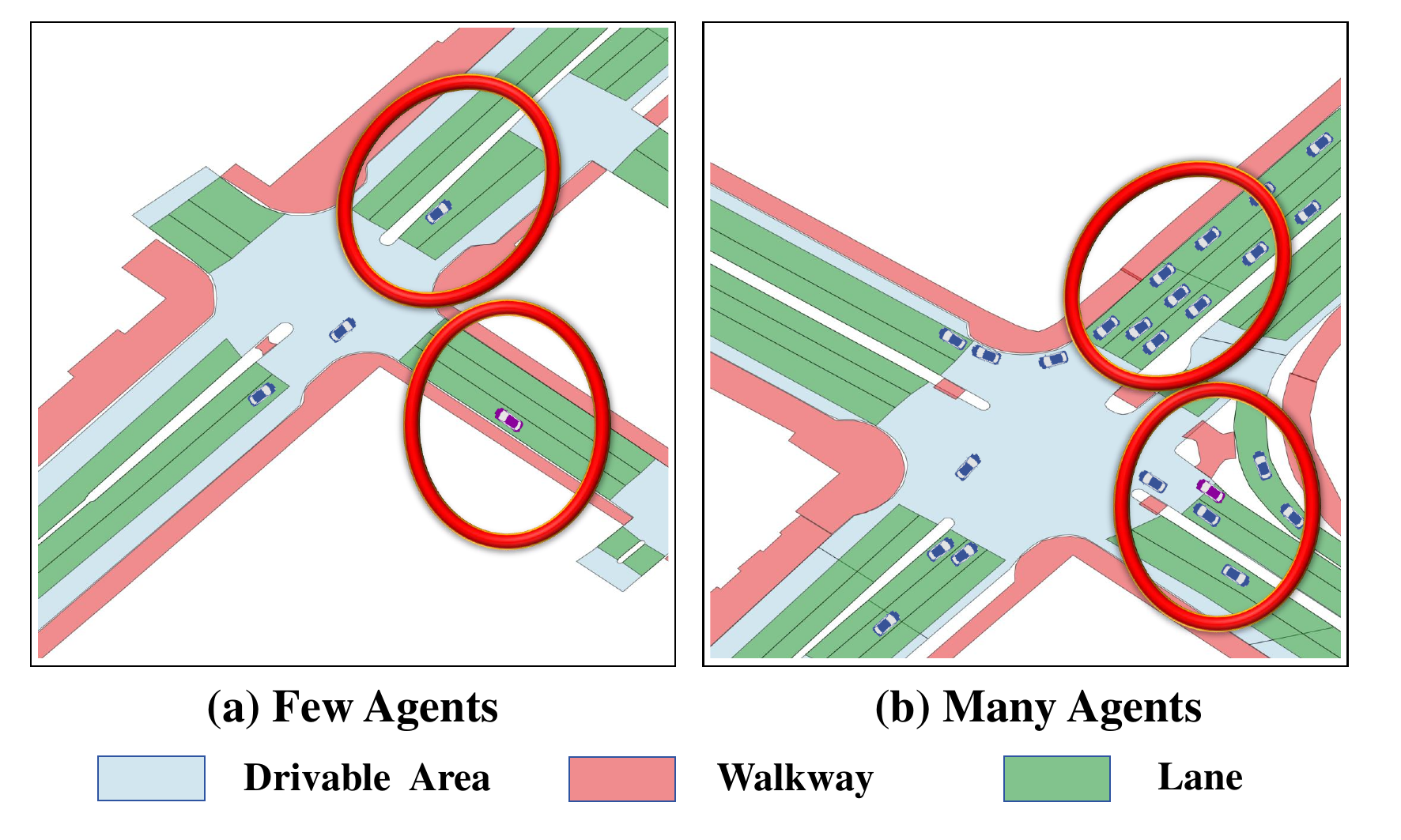}
    
    \caption{\textbf{Visualization of Trajectory Samples with different agents numbers on nuScenes.} (a) Few agents scenario. (b) Many agents scenario.}
    \label{fig:agent_density_comparison}
\end{figure}

Existing evaluation protocols primarily focus on moderate-density scenarios, which do not capture the diversity of real-world traffic. To address this, we categorize scenes into four density levels: single-agent with 1 agent, few-agent with 2 to 3 agents, medium with 4 to 8 agents, and many with 9 or more agents, and evaluate performance separately in Fig.~\ref{fig:agent_density_comparison}. This stratified setting enables a systematic analysis of how prediction accuracy changes with increasing interaction complexity.

Table~\ref{tab:mie_results} shows that higher density reduces absolute errors. For instance, weighted ADE decreases from 2.07 m in single-agent cases to 1.82 m in many-agents scenes, while weighted FDE improves from 4.67 m to 3.71 m. These decimeter-level gains are meaningful for safe driving and suggest that denser contexts should, in principle, benefit prediction.

However, MIE analysis reveals a counterintuitive trend. In many-agent scenes, MIE rises above the single-agent level, indicating that semantic maps become critical. The FDE in many-agent scenes is 4.17 m compared to 4.15 m in medium-agent scenes without maps, whereas it is reduced to 3.71 m compared to 3.88 m with maps. This paradox shows that additional agents may generate misleading signals, and semantic context is necessary to filter them, similar to how traffic rules prevent chaos at busy intersections. Conventional evaluation, which aggregates errors across all scenarios, masks this dependency. By separating semantic maps across density levels, our framework uncovers this blind spot and highlights the necessity of semantic context for robust prediction in dense traffic.

\begin{table}[tbp]
\setlength{\abovecaptionskip}{0.2cm}  
\setlength{\belowcaptionskip}{-0.1cm} 
\centering
\caption{Map Information Effectiveness across Agent Density Categories}
\label{tab:mie_results}
\setlength{\tabcolsep}{3pt} 
\footnotesize 
\begin{tabular}{lccccccc}
\hline
{Agent} & {$ADE_o$} & {$ADE_w$} & {$FDE_o$} & {$FDE_w$} & {$MIE_A$ } & \textbf{$MIE_F$ } \\
{Density} & {(m)} & {(m)} & {(m)} & {(m)} & {} & {} \\
\hline
Single (1.0)   & 2.2402 & 2.0743 & 5.1576 & 4.6749 & 0.1153 & 0.2241 \\
Few (2.5)      & 2.1612 & 2.0224 & 4.7180 & 4.2950 & 0.0972 & 0.2041 \\
Medium (5.2)   & 1.9751 & 1.8587 & 4.1527 & 3.8803 & 0.0853 & 0.1386 \\
Many (11.4)    & 1.9809 & 1.8181 & 4.1722 & 3.7113 & 0.1206 & 0.2401 \\
\hline
\end{tabular}
\end{table} 

\subsection{Spatial Distribution Comparison}
\begin{figure}[tbp]
    \centering
    \includegraphics[width=1\columnwidth]{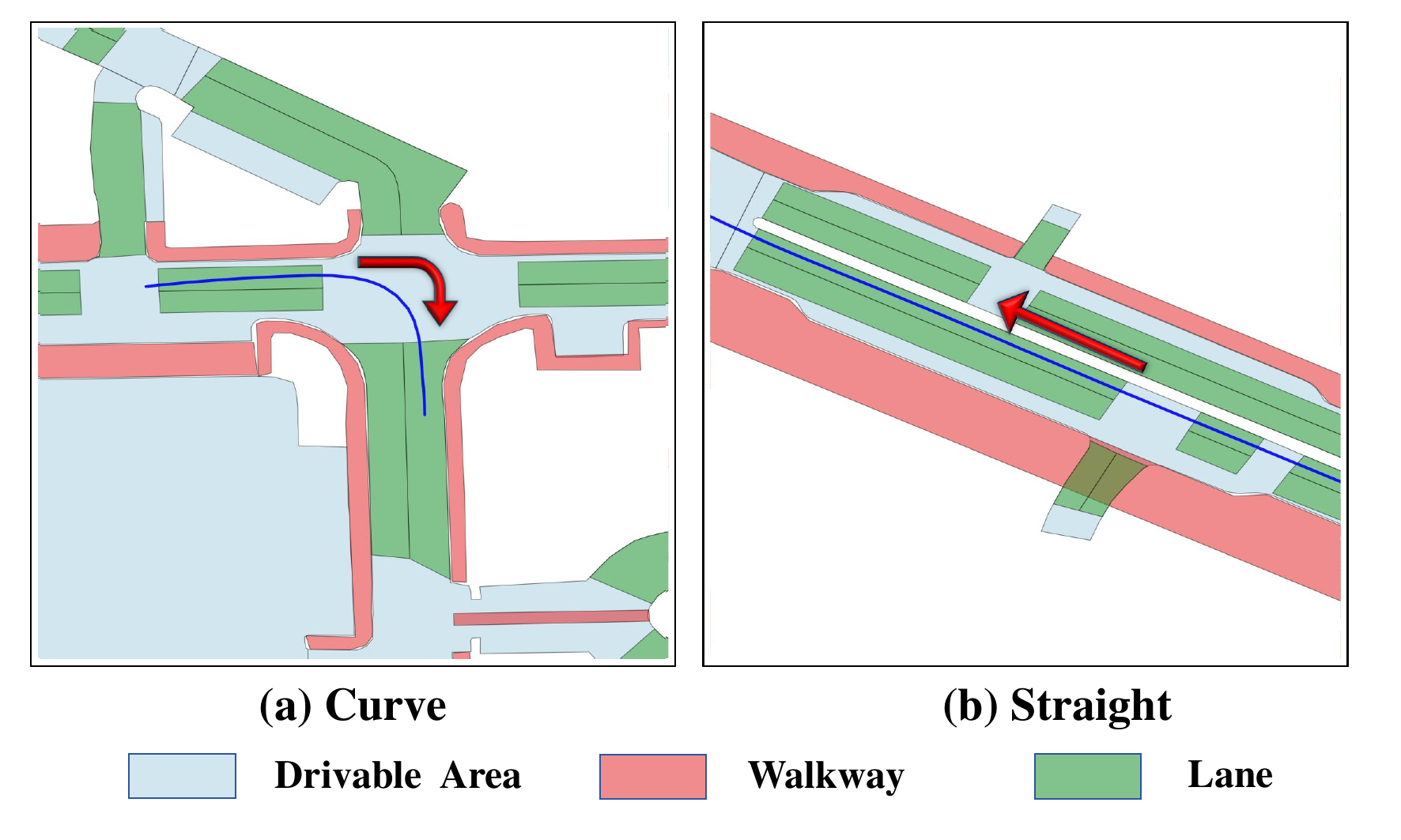}
    \caption{\textbf{Visualization of Trajectory Samples with Spatial distribution classification on nuScenes.} (a) Scenario with curve road. (b) Scenario with straight road.}
    \label{fig:spatical_distribution_compairson}
\end{figure}
Our evaluation covers various road scenarios, with curved road segments presenting greater challenges due to higher crash risks and geometric complexity. To assess model performance, we categorize the scenarios into curved and straight road segments based on semantic map processing, as shown in Fig.\ref{fig:spatical_distribution_compairson}. We evaluate prediction accuracy for both road types under two conditions: with and without semantic information. 
The results in Table~\ref{tab:spatial_mie} indicate that the absence of semantic information significantly degrades performance across both road types. For curved roads, the ADE increases from 1.95 m to 2.42 m, and the FDE rises from 4.15 m to 5.41 m when semantic information is absent. This underscores the critical role of semantic data in ensuring accurate trajectory prediction, regardless of road geometry.
When comparing road geometries, curved roads consistently show higher prediction errors than straight roads. In the absence of semantic information, curved roads exhibit an ADE of 2.42 m versus 1.95 m for straight roads, and an FDE of 5.41 m compared to 4.15 m. Even with map, the performance gap persists, with curved roads showing an ADE of 2.26 m compared to 1.82 m for straight roads. This consistent degradation underscores the inherent challenges posed by curved road geometry, which poses a significant risk to the safety of the model’s predictions.
Curved roads demonstrate a significantly higher reliance on semantic information, with values $MIE_A$ and $MIE_F$ 17. 0\% and 10. 8\%, respectively, higher than those for straight roads. The combination of curved geometry and missing semantic data represents a critical failure mode, emphasizing important safety considerations for deployment scenarios in autonomous driving.

\begin{table}[tbp]
\setlength{\abovecaptionskip}{0.2cm}  
\setlength{\belowcaptionskip}{-0.2cm} 
\centering
\caption{Map Information Effectiveness across Spatial Distribution Categories}
\label{tab:spatial_mie}
\setlength{\tabcolsep}{3pt} 
\footnotesize 
\begin{tabular}{lccccccc}
\hline
{Spacial} & {$ADE_o$} & {$ADE_w$} & {$FDE_o$} & {$FDE_w$} & {$MIE_A$ } & {$MIE_F$ } \\
{Distribution} & {(m)} & {(m)} & {(m)} & {(m)} & {} & {} \\
\hline
  Straight & 1.9493 & 1.8238 & 4.1516 & 3.8061 & 0.0929 & 0.1771 \\
  Curved   & 2.4259 & 2.2624 & 5.4093 & 4.9716 & 0.1087 & 0.1963 \\
\hline
\end{tabular}
\end{table}
\subsection{Discussion and Safety-Critical Analysis}
\begin{figure}[tbp]
\centering
\includegraphics[width=0.85\columnwidth, keepaspectratio]{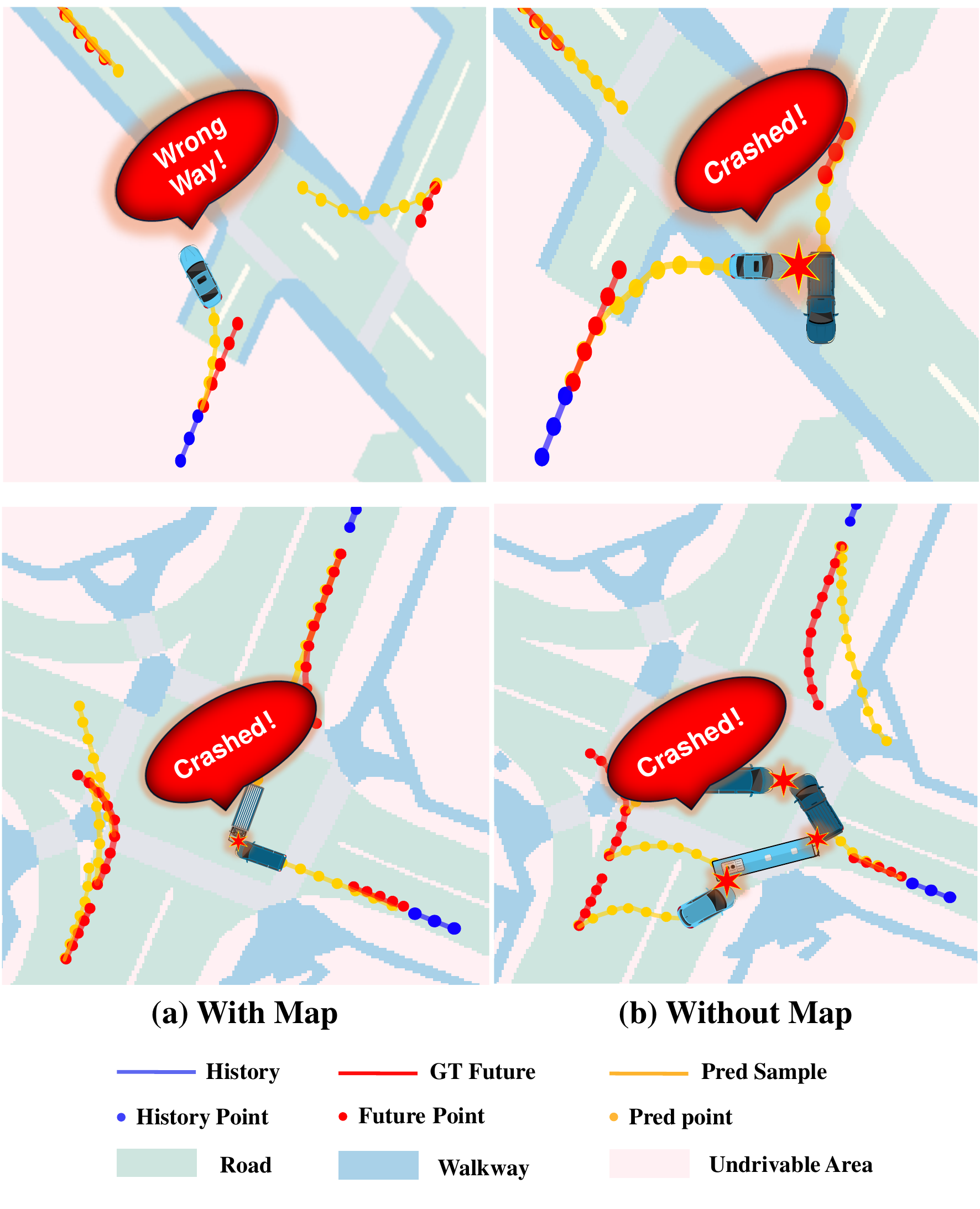}
\caption{\textbf{Visualization of Trajectory Samples with Large Errors on nuScenes.} (a) With semantic map: Despite having map information, the model still generates erroneous predictions leading to collisions. (b) Without semantic map: The absence of map information exacerbates the prediction errors, resulting in multiple collision scenarios.}
\label{fig:trajectory_comparison}
\end{figure}

Our framework addresses critical safety assessment in autonomous driving prediction. ADE/FDE provide useful overall performance scores. However, they merely focus on trajectory fitting, neglecting the impact of scene complexity on the performance of prediction models. This gap poses a safety risk in real-world driving scenarios. A model achieving acceptable aggregate metrics may fail in curved roads, without maps, or in crowded scenes. These failures can create multiple pathways to real-world accidents that current evaluation practices cannot detect.
Fig.~\ref{fig:trajectory_comparison} demonstrates our framework's effectiveness in identifying unsafe predictions that traditional metrics miss. We evaluated AgentFormer on nuScenes, focusing on two extreme scenarios (scene-0018 and scene-0301) discovered by our scenario classification system. Both cases feature complex intersections with sparse contextual agents - exactly the challenging conditions where prediction models are most vulnerable. Even with semantic maps, the model generates dangerous trajectories leading to crashes. Without maps, the failures become catastrophic with multiple collisions. While standard evaluation metrics would deem this model acceptable, our framework successfully flags these safety-critical failures, proving its value in identifying model limitations under adverse conditions.
\setlength{\belowcaptionskip}{1pt}
\section{Conclusion}
In this paper, we proposed a comprehensive evaluation framework for safety-critical prediction. Existing evaluation metrics are insufficient for selecting safe models under diverse real-world scenarios. To develop robust safety-critical prediction models, we designed a three-layer filtering framework evaluating semantic information, agent density, and spatial distribution. Our framework identifies specific risks that conventional metrics cannot expose. Through various experiments on nuScenes dataset, we demonstrated that existing state-of-the-art models exhibit significant performance degradation in challenging scenarios. Finally, our framework enables selection of truly safe autonomous driving models for real-world scenarios.
\bibliography{ifacconf}             
\end{document}